\documentclass[10pt,conference]{IEEEtran}
\IEEEoverridecommandlockouts

\usepackage[noend]{algpseudocode}
\usepackage{amsmath}
\usepackage{listings}
\usepackage{caption}
\usepackage{xcolor}
\usepackage{multirow}
\usepackage{enumitem}
\usepackage{xspace}
\usepackage{cite}
\usepackage{threeparttable}
\usepackage[linesnumbered,ruled,vlined,noend]{algorithm2e}
\usepackage{placeins}
\usepackage{float}
\usepackage{semantic}
\usepackage{mathrsfs}
\usepackage{blindtext}
\usepackage{eqnarray}
\usepackage{amssymb}
\usepackage{mathtools}
\usepackage{booktabs}
\usepackage{longtable}
\usepackage{pifont}
\usepackage{bussproofs}
\usepackage{multirow}
\usepackage{graphics}
\usepackage{wrapfig}
\usepackage{url}
\usepackage{stmaryrd}
\usepackage{multicol}

\usepackage{tikz}
\usepackage{colortbl}
\usepackage[most]{tcolorbox}

\definecolor{light-gray}{gray}{0.95}
\newcommand{\code}[1]{\colorbox{light-gray}{\texttt{#1}}}
\newcommand{\hide}[1]{}
\newcommand{\tool}{{\textsc{InvWeaver}}\xspace}
\makeatletter
\newcommand{\linebreakand}{%
  \end{@IEEEauthorhalign}
  \hfill\mbox{}\par
  \mbox{}\hfill\begin{@IEEEauthorhalign}
}
\makeatother
\newtcolorbox[auto counter, number freestyle={\noexpand\arabic{\tcbcounter}}]{definedbox}[2][]{%
    enhanced,
    colback=black!5!white,
    colframe=black!75!white,
    title=Prompt~\thetcbcounter: #2,
    fontupper=\footnotesize,
    #1
}

\begin{document}

\title{\tool: Deductive Feedback for Invariant Synthesis in Interacting-Loop Programs}

\author{
\IEEEauthorblockN{Guangyuan Wu}
\IEEEauthorblockA{\textit{Nanjing University}\\
Nanjing, China\\
guangyuanwu@smail.nju.edu.cn}
\and
\IEEEauthorblockN{Weining Cao}
\IEEEauthorblockA{\textit{Nanjing University}\\
Nanjing, China\\
weiningcao@smail.nju.edu.cn}
\and
\IEEEauthorblockN{Zehui Tan}
\IEEEauthorblockA{\textit{Nanjing University}\\
Nanjing, China\\
zehuitan@smail.nju.edu.cn}
\linebreakand
\IEEEauthorblockN{Yuan Yao}
\IEEEauthorblockA{\textit{Nanjing University}\\
Nanjing, China\\
y.yao@nju.edu.cn}
\and
\IEEEauthorblockN{Hengfeng Wei}
\IEEEauthorblockA{\textit{Hunan University}\\
Changsha, China\\
hfwei@hnu.edu.cn}
\and
\IEEEauthorblockN{Taolue Chen}
\IEEEauthorblockA{\textit{Birkbeck, University of London}\\
London, United Kingdom\\
t.chen@bbk.ac.uk}
\and
\IEEEauthorblockN{Xiaoxing Ma}
\IEEEauthorblockA{\textit{Nanjing University}\\
Nanjing, China\\
xxm@nju.edu.cn}
}

\maketitle

\begin{abstract}
Loop invariant inference is a fundamental yet challenging problem in program verification. Recent LLM-aided guess-and-check techniques have demonstrated strong performance on single-loop programs, but struggle with multi-loop programs (e.g., multiple nested and/or sequential loops), which are prevalent in classic algorithms and real-world programs. 
In this paper, we present \tool, a novel neuro-symbolic framework that synthesizes invariants for multi-loop programs. The key insight is to explicitly expose inter-loop dependencies and systematically propagate proof obligations through a principled combination of abstraction and deductive reasoning. 
%
%
%
%
We evaluate \tool on a comprehensive benchmark suite, including a newly curated dataset derived from classic algorithms. Experimental results show that \tool substantially outperforms the existing methods. In particular, \tool solves 72 out of 82 multi-loop benchmark problems, exceeding the strongest competitor by 32 problems, while maintaining superior performance on single-loop tasks. 
\end{abstract}


\section{Introduction}\label{sec:intro}
Loop invariants are central to program verification, serving as the inductive backbone for proving correctness and safety properties. A loop invariant is a property that holds before and after every iteration of a loop, forming the inductive bridge in Hoare-style reasoning and related verification frameworks. However, automatically inferring such invariants remains a long-standing challenge, due to the inherent undecidability of the problem and the complexity of real-world programs.

\hide{
\smallskip
{\bf Existing work}.
Early research on invariant inference was dominated by deductive reasoning approaches, which relied on techniques such as constraint solving~\cite{colon2003linear,gupta2009tests,hojjat2018eldarica,vediramana2024global}, abstract interpretation~\cite{cousot1979systematic, cousot1978automatic, karr1976affine}, and Craig interpolation~\cite{jhala2006practical,mcmillan2010lazy}. While these methods provide strong soundness guarantees, they often struggle with scalability issues (e.g., state-space explosion) and require significant manual intervention, particularly for non-linear arithmetic programs.
To overcome these limitations, inductive and template-based approaches emerged~\cite{ernst2007daikon,sharma2013data,nguyen2014dig}. 
They restrict the search space to parametric forms—typically polynomial templates—and use algebraic solvers to instantiate coefficients. However, the reliance on templates introduces a inherent trade-off: simple templates lack expressiveness, whereas complex ones quickly lead to intractable constraints. Similarly, early data-driven methods~\cite{zhu2018data,freund1998large} that framed invariant inference as a supervised learning task (e.g., with SVMs) suffered from heavy feature engineering and poor generalization.
}

Recently, the field has adopted 
a guess-and-check framework, where candidate invariants are iteratively generated and verified. Although heuristic and incomplete, this framework offers superior scalability and flexibility by allowing data-driven inference to complement symbolic reasoning. Diverse learning techniques have been instantiated in this framework, including decision trees~\cite{ezudheen2018horn,garg2016learning,xu2020interval}, reinforcement learning~\cite{si2018learning,yu2023loop}, continuous logic networks~\cite{ryan2019cln2inv,yao2020learning}, and, most recently, large language models (LLMs)~\cite{kamath2023finding,chakraborty2023ranking,liu2024towards,wu2023lemur}.

Specifically, the LLM-aided guess-and-check methods have shown remarkable performance for programs containing a single loop. However, real-world programs frequently feature complex (multiple nested and/or sequential) loop structures, where computations must iterate over multiple dimensions, stages, or interacting entities. Some representative examples include array and matrix processing, sorting algorithms, graph algorithms, and dynamic programming, to name a few. When applied to such complex loops, existing methods often suffer from \emph{local under-specification}.
Namely, inner or intermediate loops typically lack explicit contextual information (e.g., precise pre-conditions or post-conditions) during reasoning, causing LLMs to hallucinate ``weak invariants'' that are syntactically valid but fail to capture the essential semantic dependencies between adjacent loops. As a result, these invariants are insufficient to support global correctness proofs, even though they may appear locally plausible (see Section~\ref{sec:motiv} for an example, where existing methods fail even on the {\em bubble sort} algorithm).

\smallskip
\noindent{\bf Our work}.
%
To address the local under-specification issue, we propose \tool, a neuro-symbolic framework for synthesizing loop invariants in multi-loop programs. 
\tool is built upon a \textit{loop-level call graph} (LCG), which is a topology specifically designed to record and expose inter-dependencies among all the loops in the current program; it then orchestrates an obligation-guided, context-aware inference process.
Specifically, on the neural side, \tool adopts an LLM-based bi-directional and top-down inference strategy tailored to multiple loops.
By leveraging the LCG, LLMs are supplied with a structured verification context for each loop, enabling them to reason beyond local loop semantics and to precisely capture the global constraints propagated from predecessor and successor loops. 
On the symbolic side, in contrast to existing single-loop based verification that provides feedback for current loop only, we introduce a {\em WP-guided refinement} mechanism to 
{\em propagate} proof obligations across the loop boundaries. Specifically, \tool extracts WP-derived verification conditions as explicit repair targets and propagates them over the LCG. Grounded in the weakest precondition (WP) calculus as a rigorous deductive basis, this feedback loop prompts the LLM to strengthen or weaken generated invariants in response to localized proof failures.
As a result, invariants associated with adjacent loops are jointly refined, ensuring their convergence to a consistent inductive proof chain across the loop hierarchy.

Furthermore, to enhance robustness, we implement a delayed filtering mechanism that retains partially valid invariants, discarding them only after they are proven invalid in multiple verification contexts. This prevents the premature elimination of useful semantic information and enables the reuse of valuable intermediate reasoning. 

We evaluate \tool on a diverse suite of loop invariant inference benchmarks. 
In addition to traditional numerical programs from SV-COMP~\cite{beyer2022progress} and OOPSLA~\cite{dillig2013inductive}, we introduce a new dataset derived from the highly influential textbook \textbf{Introduction to Algorithms} (CLRS)~\cite{cormen2022introduction}. This dataset features algorithms over 
various complex loop structures, frequently requiring quantified invariants, and serving as the building blocks for real-world programs. 
We compare \tool against with 7 representative state-of-the-art methods, including UAutomizer~\cite{heizmann2013software}, Code2Inv~\cite{si2018learning}, CLN2INV~\cite{ryan2019cln2inv}, G-CLN~\cite{yao2020learning}, LEMUR~\cite{wu2023lemur}, LaM4Inv~\cite{wu2024llm}, and AutoSpec~\cite{wen2024enchanting}. 
Experimental results demonstrate that \tool successfully solves 72 out of 82 multi-loop benchmark problems, which is at least 32 instances more than the existing competitors. 

\smallskip
\noindent{\bf Contribution}.
The main contributions of this paper include: 
\begin{itemize}[leftmargin=*] 
    \item \textit{Approach.} We propose \tool, a novel neuro-symbolic loop invariant inference tool designed for multi-loop programs. By integrating an hierarchical, context-aware inference strategy with a WP-guided refinement mechanism, \tool bridges the contextual gap in multi-loop reasoning, enabling the synthesis of expressive invariants that capture inter-loop dependencies.

    \item \textit{Dataset and Evaluation.} We extend the existing benchmark landscape by curating a new dataset, tripling the number of loop invariant inference tasks involving multiple loops. The dataset covers a diverse set of classic algorithmic categories, including searching, sorting, dynamic programming, and graph algorithms. Extensive experiments on this benchmark, as well as on existing benchmarks, demonstrate that \tool consistently advances the state-of-the-art. 
\end{itemize}

\smallskip
\noindent{\em Roadmap.} The rest of the paper is organized as follows. Section~\ref{sec:pre_mot} provides the necessary background. Section~\ref{sec:motiv} presents a motivating example. Section~\ref{sec:appro} describes the proposed approach, and Section~\ref{sec:exper} reports the evaluation results. Section~\ref{sec:discu} discusses limitations and threats to validity, and Section~\ref{sec:relate} reviews the related work. Section~\ref{sec:concl} concludes.

\section{Preliminary}\label{sec:pre_mot}


\subsection{Loop Invariant Inference}\label{sec:problem}
Loop invariant inference is a central problem in program verification, aiming to automatically synthesize inductive assertions that enable the proof of correctness for iterative constructs. 
Formally, consider a Hoare triple of the form \code{\{$P$\} while $B$ do $S$ \{$Q$\}}, where $P$ denotes the pre-condition that holds before entering the loop, $B$ is the loop guard, $S$ is the loop body, and $Q$ is the post-condition that is expected to hold upon loop termination.

According to classical Hoare logic, the correctness of this loop can be established by identifying a loop invariant $I$ such that the following rule holds:
\[\label{eq:invariant_properties}
\inference{P \implies I \quad \{I \land B\} \, S \, \{I\} \quad (I \land \neg B) \implies Q}
{\{P\} \; \text{while} \; B \; \text{do} \; S \; \{Q\}}.
\]
The above rule imposes three requirements on a valid loop invariant:
\begin{itemize}[leftmargin=*]
\item \textit{Establishment}. The invariant must hold before the first iteration of the loop, i.e., the pre-condition $P$ implies the invariant $I$. This ensures that the invariant is initially established upon loop entry.
\item \textit{Preservation}. Assuming the invariant holds at the beginning of a loop iteration and the loop guard $B$ evaluates to true, executing the loop body $S$ must preserve the invariant. That is, the invariant remains true after each iteration of the loop.
\item \textit{Provability}. When the loop terminates, the invariant together with the negation of the loop guard must imply the desired post-condition $Q$. This ensures that the invariant is sufficiently strong to establish the correctness of the program after the loop completes.
\end{itemize}
In practice, SMT solvers such as Z3~\cite{de2008z3} are commonly used to discharge these three proof obligations. If any condition fails, the solver provides a counterexample indicating the source of failure. 
However, synthesizing such an invariant 
is highly non-trivial and, in general, undecidable. As a result, practical approaches rely on incomplete and heuristic techniques that iteratively hypothesize, validate, and refine candidate invariants through solver-guided feedback.

\subsection{Weakest Precondition Calculus}\label{sec:wp}
The weakest precondition (WP) calculus~\cite{dijkstra1976discipline}, introduced by Dijkstra, provides a rigorous basis for defining program semantics through predicate transformers. Given a program statement $S$ and a post-condition $Q$, the weakest precondition $wp(S, Q)$ characterizes the set of all initial states from which the execution of $S$ is guaranteed to terminate in a state satisfying $Q$. By construction, $wp(S,Q)$ is the weakest (i.e., least restrictive) such precondition that ensures this guarantee.

The WP calculus is defined inductively over the syntactic structure of program statements. In particular, the semantics for assignment, sequence, assumption, and assertion can be  formalized as 
    \begin{align*}
        wp(x := e, Q) & \equiv Q[e/x]  \\   wp(S_1; S_2, Q) & \equiv wp(S_1, wp(S_2, Q)) \\
        wp(\textbf{assume } P, Q) & \equiv P \implies Q \\ 
        wp(\textbf{assert } P, Q) & \equiv P \land Q   
    \end{align*}
where $Q[e/x]$ denotes the capture-avoiding substitution of variable $x$ with expression $e$ in predicate $Q$. The sequencing rule dictates a strictly backward propagation of logical constraints from the post-state to the pre-state. The \textbf{assume} statement restricts the feasible execution paths, namely, if the condition $P$ is violated, the corresponding execution path is discarded. The \textbf{assert} statement imposes a strict verification obligation, requiring $P$ to hold to prevent verification failure.


\section{Motivation}\label{sec:motiv}
This section presents a motivating example. Specifically,
Listing~\ref{lst:bubble_motivating} shows a C implementation of the classic {\em bubble sort} algorithm, with specifications and loop invariants written in ANSI/ISO C Specification Language (ACSL)~\cite{cuoq2012frama}. ACSL is a behavioral interface specification language (BISL)~\cite{hatcliff2012behavioral} in the Frama-C framework for formally specifying behavioral properties of C source code. The post-condition specifies that the array $a$ must be sorted in ascending order: 
$ \forall i, j \in [0, \text{length}), i < j \Rightarrow a[i] \leq a[j]$.
To formally verify this property, a collection of loop invariant clauses needs to be synthesized (i.e., from $I_0$ to $I_8$), as annotated in Listing~\ref{lst:bubble_motivating}.

Despite the simplicity of bubble sort, existing tools (e.g., LEMUR~\cite{wu2023lemur}, AutoSpec~\cite{wen2024enchanting}, and LaM4Inv~\cite{wu2024llm}) struggle to generate the complete loop invariant. Specifically, they fail to generate critical clauses such as $I_7$ and $I_8$, which is not necessarily useful for verifying the inner loop alone, but essential to bridge the reasoning chain for verifying the outer loop and the entire program. 
In particular, the {\em preservation} of the outer-loop invariant $I_2$ is coupled with the {\em provability} of the inner-loop invariants, which must be strong enough to entail the post-loop summary required by the enclosing loop. 
As a result, while $I_2$ is semantically valid, the absence of sufficiently strong 
inner-loop invariants (e.g., the lack of $I_7$ and $I_8$) breaks the Hoare-logic reasoning chain. 
This example highlights a fundamental limitation in existing methods: they formulate loop invariant refinement as a local repair problem for the failing loop, and use verification feedback only as a rejection signal, rather than deriving the missing inter-loop proof obligations as explicit synthesis targets.

\begin{lstlisting}[
float=!t,
language=C,
caption={[The bubble sort example from our benchmark.]The bubble sort example from our benchmark.\protect\rule[-9pt]{0pt}{0pt}},
captionpos=t,
keywordstyle=\bfseries\color{blue},
commentstyle=\color{green!60!black},
basicstyle=\ttfamily\scriptsize,
numbers=left,
numberstyle=\scriptsize,
numbersep=5pt,
stepnumber=1,
frame=single,
xleftmargin=1.5em,
framexleftmargin=2em,
breaklines=true,
breakatwhitespace=false,
columns=fullflexible,
keepspaces=true,
escapeinside={(*@}{@*)},
label={lst:bubble_motivating}
]

/*@
predicate sorted(int* a, integer lo, integer hi) =
\forall integer i, j; lo <= i <= j < hi ==> a[i] <= a[j];
*/
/*@
requires 1 < length;
requires \valid(a+(0..length-1));
assigns a[0..length-1];
ensures e_1: sorted(a, 0, length);
*/
void _bubble_sort(int *a, int length) {
    int up = 1;
    int down;
    /*@
    loop invariant I_0: 1 < length;
    loop invariant I_1: 1 <= up <= length;
    loop invariant I_2: sorted(a, 0, up);
    loop assigns down, up, a[0..length - 1];
    */
    for (; up < length; up++) { 
        down = up;
        /*@
        loop invariant I_3: 1 < length;
        loop invariant I_4: 1 <= up < length;
        loop invariant I_5: 0 <= down <= up;
        loop invariant I_6: sorted(a, 0, down);
        loop invariant I_7: sorted(a, down, up + 1);
        loop invariant I_8: \forall integer k1, k2; 
            0 <= k1 < down < k2 <= up ==> a[k1] <= a[k2];
        loop assigns down, a[0..length - 1];
        */
        while (0 < down && a[down] < a[down - 1]) { 
            int tmp = a[down];
            a[down] = a[down - 1];
            a[down - 1] = tmp;
            down = down - 1;
        }
    }
}
\end{lstlisting}

To better understand this example, we illustrate how a human verifier would derive the key clauses. Since the outer loop concludes with the statement \texttt{up++}, the post-condition \texttt{sorted(a, 0, up + 1)} must hold after the inner loop terminates. However, the existing inner-loop invariant $I_6$ only guarantees order over the lower segment \texttt{[0, down)}, which is insufficient to reestablish the outer invariant.
Examining the inner-loop guard \texttt{(0 < down $\&\&$ a[down] < a[down-1])} reveals that, upon exit, the negation \texttt{(down <= 0 || a[down] >= a[down-1])} holds. Combining this exit condition with the current invariant implies that the sortedness property can be extended to cover the range \texttt{[down, up + 1)}, leading to the introduction of clause $I_7$.
Re-evaluating the proof obligations, however, shows that this extension causes a {\em preservation} failure within the inner loop itself. By applying backward reasoning to the Hoare triple $\{I \land B\} S \{I\}$, we identify that, after one loop iteration, it must be shown that \texttt{a[down - 1]} is less than all elements in the range \texttt{(down, up + 1)} following the swap operation. This condition ensures that the ordering property propagates across iterations and corresponds precisely to invariant $I_8$ after inductive generalization.

This reasoning process reveals two methodological insights. First, the correctness of each invariant depends hierarchically on others. To model this interdependence, we introduce the loop-level call graph (LCG), an abstraction that captures control dependencies among loops. Second, the missing clauses ($I_7$ and $I_8$) directly correspond to the weakest preconditions necessary to reestablish the outer invariant $I_2$. The WP calculus thus provides a formal mechanism to compute these logical targets. 
%
These two insights motivate our proposed framework, which leverages program abstraction and WP calculus to propagate proof obligations across loop boundaries. 
%

\begin{figure*}[t]
  \centering
  \includegraphics[width=0.94\textwidth]{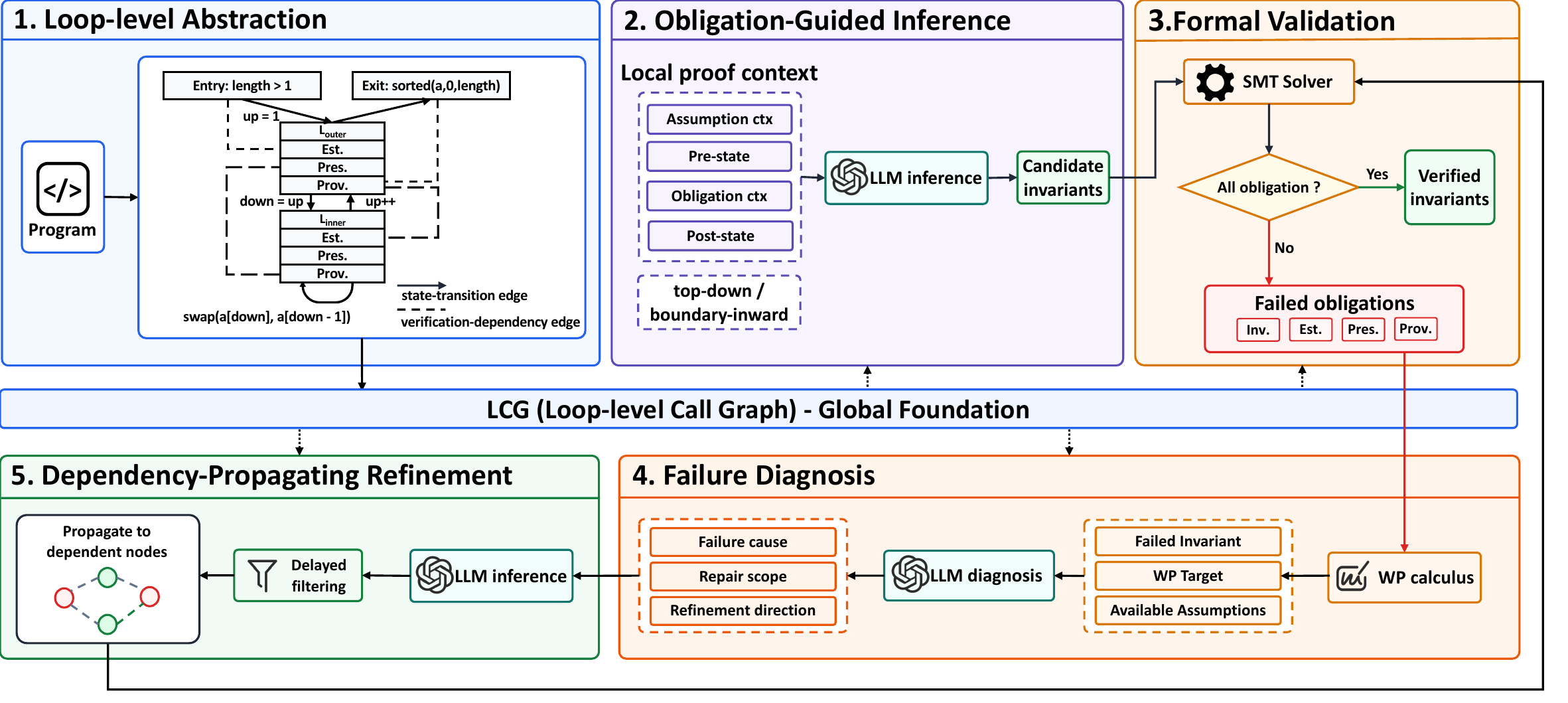}
  \caption{Overview of \tool}
  \label{fig:overview}
\end{figure*}
\section{Approach}\label{sec:appro}
In this section, we present \tool, which integrates the weakest precondition calculus with LLMs to automate loop invariant inference for multi-loop programs. 
Figure~\ref{fig:overview} presents the overall architecture of our approach, which consists of the following five iterative steps.

\begin{enumerate}[leftmargin=*]
\item \textbf{Loop-level abstraction}. \tool first builds an LCG as the foundation of the entire approach. The nodes of the LCG are loops or boundary contexts, and 
edges encode inter-loop state transfer and proof-obligation dependencies.
\item \textbf{Obligation-guided inference}. \tool then traverses the LCG top-down for nested loops and boundary-inward for sequential loops, using dependency-relevant specifications as local proof contexts to prompt initial invariants.
\item \textbf{Formal validation}. A verifier checks the three properties of establishment, preservation, and provability, either passing the validation or returning the failed invariants and violated obligations when the verification fails.
\item \textbf{Failure diagnosis}. For each failed obligation, WP calculus reconstructs the proof targets, while the LLM identifies the failure cause, modification scope, and refinement direction.
\item \textbf{Dependency-propagating refinement}. Guided by the diagnosis, \tool repairs the failed and dependent invariants, applies delayed filtering to discard irrecoverable candidates, and propagates newly exposed obligations along the LCG. It then returns Step 3. 
\end{enumerate}

\subsection{Loop-level Abstraction}
The effectiveness of LLMs in program verification is often hindered by the complexity of raw source code and, in particular, the tight coupling of 
logic across multiple loops. To address this, we introduce a loop-level abstraction that transforms the raw program text into a structured representation termed the loop-level call graph (LCG).
The LCG abstracts the program into loop-level entities and distinguishes two kinds of inter-loop relations: state-transition relations induced by executable control flow, and verification-dependency relations induced by proof obligations among loop invariants.
This abstraction elevates the reasoning granularity from statement-level syntax to loop-level semantics, explicitly modeling each loop together with its surrounding boundary conditions as distinct logical entities.
By exposing both structural nesting and proof-obligation dependencies, this representation provides LLMs with a concise verification context for invariant inference across nested and interacting control structures.

The LCG abstraction algorithm adopts a hybrid static analysis approach that traverses both the abstract syntax tree (AST) and the control flow graph (CFG) of the target program $P$.
Specifically, an LCG is defined as a tuple $G = (V, E_{st}, E_{vd})$, where the node set $V$ is partitioned into two disjoint subsets $V_{bound}$ and $V_{loop}$. $V_{bound}$ represents the program entry and exit points, while $V_{loop}$ represents individual loop constructs (e.g., \texttt{for/while/do-while}).
The state-transition edge set $E_{st}$ captures executable transfers between nodes. Each edge $e = (u, v) \in E_{st}$ is annotated with the code segments executed during the transition from node $u$ to node $v$, including initialization statements, updates between nested iterations, and loop-body statements.
The verification-dependency edge set $E_{vd}$ captures logical dependencies between loop invariants. Each edge in $E_{vd}$ records that discharging an establishment, preservation, or provability obligation for one invariant requires the specification supplied by another loop or boundary node.
\tool constructs $E_{vd}$ as an obligation-labeled dependency graph.
For a state-transition edge $(u,v) \in E_{st}$, \tool binds the provability obligation of predecessor $u$ to the establishment obligation of successor $v$; for a nested pair $(u,v)$ where $u$ is an enclosing loop and $v$ is the terminal inner loop reaching its back-edge, \tool binds the provability obligation of $v$ to the preservation obligation of $u$.
Figure~\ref{fig:overview} illustrates the LCG abstraction for a representative nested-loop structure, where loop nodes, boundary contexts, state-transition edges, and verification-dependency edges jointly expose the verification context propagated in later phases.

In subsequent phases, \tool uses the LCG as the common substrate for obligation-guided inference, formal validation, failure diagnosis, and dependency-propagating refinement.
State-transition edges provide executable context for the current loop, while verification-dependency edges supply the local proof contexts used for prompting and determine which invariants should be jointly repaired when verifier feedback exposes a missing proof obligation.
This enables the model to reason about both local properties (e.g., partial ordering within a prefix) and global correctness (e.g., full array sorting) in a precise and compositional way.

\subsection{Obligation-Guided Inference}
Our obligation-guided inference mechanism comprises two key components: (i) a traversal strategy which schedules inference tasks based on LCG, and (ii) a prompt template designed to support proof-context-aware LLM generation.

\subsubsection{Inference Strategy}
Based on the LCG constructed in the previous phase, we schedule invariant inference to maximize the contextual information available for candidate generation.
Unlike traditional single-loop invariant generation, where the global pre-condition and post-condition can be directly used as local constraints, programs with multiple loops often leave each individual loop under-specified.
To mitigate this problem, \tool adopts a hierarchical traversal strategy: for nested loops, it proceeds top-down from the outermost loop to the innermost loop; for sequential loops at the same nesting depth, it performs boundary-inward traversal from the first and last loops toward the intermediate loops.
During this traversal, establishment-dependent specifications, including loop invariants and global contracts, form the local pre-condition, while provability-dependent specifications form the local post-condition.
As inference proceeds, each newly generated invariant is recorded and becomes available to dependent nodes through $E_{vd}$.

The rationale behind this traversal strategy is to maximize the contextual constraints available to the LLM during loop invariant inference, which is pivotal for guiding LLM-based generation.
Empirically, we observe that without these propagated constraints, LLMs frequently hallucinate invariants that are locally valid yet deductively insufficient to establish the overall proof.
For example, in the sorting algorithm in Listing~\ref{lst:bubble_motivating}, without the knowledge that the outer loop partitions the array, the model may generate redundant predicates such as
\texttt{sorted(a, 0, down)}, while failing to capture the essential relational invariant between partitions.
By supplying explicit constraints along the LCG's verification-dependency edges, we can 
steer LLMs towards invariants that can logically bridge 
adjacent loop structures.



\subsubsection{Obligation-Guided Prompt}

{\small
\begin{figure}[t]
\begin{definedbox}[label=prompt:ObligationGuidedInference]{Obligation-Guided Inference Prompt.}
    \texttt{\{PROGRAM\}} \\
    You are an expert in program verification. Please generate initial loop invariant candidates as C annotation comments at the hint location (annotated by ``[Hint] Please infer the loop invariant for the following loop.'') using ACSL language strictly based on the program's execution logic and standard correctness. ACSL is a specification language for C programs that conforms to the design by contract paradigm, utilizing Hoare style pre- and postconditions and invariants.  \\
    Specifically, for the current verification task:
    \begin{itemize}[leftmargin=1.5em, nosep]
        \item $\{\texttt{ASSUMPTION\_CONTEXT}\}\ \texttt{PRE\_STATE}\ \{I\}$: after the pre-state transition, the assumption context should establish the generated invariant.
        \item $\{\texttt{I} \wedge \neg \texttt{LOOP\_CONDITION}\}\ \texttt{POST\_STATE}$\\
        $\{\texttt{OBLIGATION\_CONTEXT}\}$: when the current loop exits, the generated invariant together with the negated loop condition should entail the obligation context after the post-state transition.
    \end{itemize}
    Please focus on invariants justified by the reachability and provability contexts above, and verify that the generated loop invariant is consistent with these proof contexts. Preservation is not the current focus and will be handled in later steps.\\
    Use \texttt{\detokenize{&&}}, \texttt{\detokenize{||}}, \texttt{\detokenize{==>}}, \texttt{\detokenize{\forall}}, or \texttt{\detokenize{\exists}} if necessary. 
    Your answer should follow the following format:
    \begin{tcolorbox}[colback=white, boxrule=0.2pt, arc=1pt, top=1mm, bottom=1mm]
        \small\ttfamily
        loop invariant ...;\\
        ...
    \end{tcolorbox}
    No explanation. No commentary. Just show me the loop invariant.
\end{definedbox}
\vspace{-5mm}
\end{figure}
}

Prompt~\ref{prompt:ObligationGuidedInference} presents the template used in obligation-guided inference. We first instantiate the verification context by replacing the placeholder ``\texttt{{PROGRAM}}'' with the concrete program code.
To align the model's generation domain with formal verification standards, we define its role as an expert in program verification and specify the target language as ACSL (cf. Section~\ref{sec:motiv}).
The fields ``\texttt{{ASSUMPTION\_CONTEXT}}'' and ``\texttt{{OBLIGATION\_CONTEXT}}'' are instantiated with dependency-relevant specifications from loop or boundary nodes along $E_{vd}$, while ``\texttt{{PRE\_STATE}}'' and ``\texttt{{POST\_STATE}}'' are extracted from state-transition edge set $E_{st}$.
Together, they encode the local Hoare-style obligations $\{P\}\ S_{\mathit{pre}}\ \{I\}$ and $\{I \wedge \neg B\}\ S_{\mathit{post}}\ \{Q\}$.
Accordingly, the prompt frames initial invariant generation as a boundary-alignment problem: synthesize predicates that are reachable from the local entry context and strong enough to discharge the local exit obligation.
The inductive preservation obligation is deferred to validation and refinement. In nested loops, preservation of an enclosing invariant depends on modular summaries supplied by inner-loop invariants; before such summaries are available, enforcing $\{I \wedge B\}\ S_{\mathit{body}}\ \{I\}$ can over-constrain generation and lead to speculative strengthening. The subsequent failure diagnosis and dependency-propagating refinement phases use WP feedback to decide whether to relax the current invariant or strengthen a dependent loop summary.
We conclude the prompt with strict formatting instructions requiring the model to 
omit conversational filler and output only the invariant candidates for subsequent verification.

\subsection{Failure Diagnosis}
{\small
\begin{figure}[t]
\begin{definedbox}[label=prompt:failure-diagnosis]{Failure Diagnosis Prompt.}
    \raggedright
    \texttt{\{PROGRAM\}} \\
    You are an expert in program verification using ACSL and Frama-C. Diagnose why the following verification obligation cannot be discharged. Do not propose new invariants.\\
    {[Failed Invariant]} \texttt{\{ERROR\_INV\}} \\
    {[WP Target]} \texttt{\{WP\_GOAL\}} \\
    {[Available Assumptions]} \texttt{\{INNER\_INV\}} and $\neg$\texttt{\{INNER\_LOOP\_CONDITION\}} \\
    Analyze the proof failure by answering the following items:
    \begin{itemize}[leftmargin=1.5em, nosep]
        \item Whether failed invariant itself is too strong or non-inductive with respect to the loop body.
        \item Whether the invariant(s) of \texttt{\{INNER\_LOOP\}} are too weak to imply the WP goal after the inner loop terminates.
        \item Which missing logical fact is needed to bridge the available assumptions and WP target.
    \end{itemize}

    IMPORTANT RULES:
    \begin{enumerate}[leftmargin=1.5em, nosep, label=\arabic*.]
        \item Output only a diagnosis object using the following fields:
        \begin{tcolorbox}[colback=white, boxrule=0.2pt, arc=1pt, top=1mm, bottom=1mm]
            \small\ttfamily
            {} [Failure Cause]\\
            ...\\
            {} [Responsible Scope]\\
            ...\\
            {} [Refinement Direction]\\
            ...
        \end{tcolorbox}
        \item The responsible scope must be selected from \texttt{\{CUR\_LOOP\}}, \texttt{\{INNER\_LOOP\}}, or both.
        \item Boundary contracts may be cited as assumptions or targets, but must not be selected as repair targets.
    \end{enumerate}
\end{definedbox}
\vspace{-5mm}
\end{figure}
}

Following the obligation-guided inference phase, the generated candidate invariants are first checked by an SMT-backed verifier, which attempts to discharge their establishment, preservation, and provability obligations.
If verification fails, the verifier returns a set of failed proof obligations, each associated with an invariant and an obligation type.
For each failed obligation, \tool invokes WP calculus to reconstruct the corresponding verification condition and derive logical bounds for diagnosis.
We employ WP calculus as the deductive backbone of our framework due to its ability to precisely capture the logical boundaries within which invariant refinement must operate.
Existing LLM-based invariant inference methods typically use verification feedback as a rejection signal or a local error message, which indicates that a candidate fails but provides limited guidance on where the proof gap should be repaired.
In contrast, \tool reifies WP feedback into structured refinement targets: by computing $wp(S, Q)$, it derives the weakest condition over the pre-state that guarantees the post-condition $Q$ after executing statement $S$.
This condition exposes the target that a repaired invariant must entail or be entailed by, thereby determining whether the current candidate should be strengthened or weakened.
The resulting WP-derived target is then associated with the corresponding LCG dependency slice, allowing proof obligations extracted from the WP plugin to be propagated beyond the locally failing loop.

The failed obligation remains the unit of diagnosis. However, its diagnostic context is dependency-sliced.
If the corresponding obligation is incident to verification-dependency edges in the LCG, \tool includes the specifications of the related loop nodes in the same proof context and marks them as potential repair locations.
This is necessary because an unprovable obligation may be only the symptom of a proof gap in a dependent loop summary.
For instance, in the motivating example in Listing~\ref{lst:bubble_motivating}, the failed verification condition appears as a preservation failure of the outer-loop invariant, but the missing fact is supplied by the provability of the terminal inner loop: the inner-loop invariant must be strong enough to entail the post-loop summary required by the enclosing loop.
Diagnosing the outer invariant alone would therefore mislocalize the repair target, whereas the dependency slice exposes both the failing obligation and the proof-relevant inner-loop summary.
Entry and exit nodes are treated as trusted boundary contracts: they may contribute assumptions or proof targets, but are not selected as repair targets.
Prompt~\ref{prompt:failure-diagnosis} shows the generic template. It asks the LLM to compare the failed invariant with the WP-derived target under the dependency-sliced proof context, and to report the failure cause, responsible repair location, and refinement direction.

\subsection{Dependency-Propagating Refinement}
Dependency-propagating refinement treats invariant repair as an iterative propagation of proof obligations over the LCG.
For each failed verification obligation, \tool instantiates the prompt with the failed invariant and the diagnosis result, including the failure cause, repair scope, and refinement direction.
The generated update is then revalidated by the verifier; if it exposes another failed obligation, \tool diagnoses the new failure and propagates the repair to the next proof-relevant loop node.
Thus, refinement is not a one-shot local patch, but a feedback loop that moves along verification-dependency edges until the affected proof chain is closed.
The motivating example illustrates this propagation.
To reestablish the outer-loop invariant, \tool first derives the need for $I_7$, which summarizes the sorted segment after the inner loop terminates.
However, adding $I_7$ introduces a new preservation obligation inside the inner loop.
The next WP-guided diagnosis propagates the repair inward and derives the missing ordering fact that becomes $I_8$ after inductive generalization.
In this way, the repair of one invariant may expose a downstream obligation whose resolution requires another refinement step.
Prompt~\ref{prompt:dependency-refinement} shows the template used for each propagation step.

{\small
\begin{figure}[t]
\begin{definedbox}[label=prompt:dependency-refinement,halign title=left]{Dependency-Propagating\\Refinement Prompt.}
    \raggedright
    \texttt{\{PROGRAM\}} \\
    You are an expert in program verification using ACSL and Frama-C. Refine only the loop invariants in the repair scope so that the failed verification obligation can be discharged.\\
    {[Failed Invariant]} \texttt{\{ERROR\_INV\}} \\
    {[Failure Cause]} \texttt{\{FAILURE\_CAUSE\}} \\
    {[Repair Scope]} \texttt{\{RESPONSIBLE\_SCOPE\}} \\
    {[Refinement Direction]} \texttt{\{REFINEMENT\_DIRECTION\}} \\
    Use the diagnosis to refine the failed invariant and its related invariants.\\
    IMPORTANT RULES:
    \begin{enumerate}[leftmargin=1.5em, nosep, label=\arabic*.]
        \item Modify only invariants in \texttt{\{RESPONSIBLE\_SCOPE\}}.
        \item Your answer should follow the following format:
        \begin{tcolorbox}[colback=white, boxrule=0.2pt, arc=1pt, top=1mm, bottom=1mm]
            \small\ttfamily
            [Loop \texttt{\{LOOP\_IN\_SCOPE\}}]\\
            loop invariant ...;\\
            ...;
        \end{tcolorbox}
        Output one block for each loop in the repair scope. All invariants must strictly satisfy ACSL rules.
        \item Do not output any of the original invariants. Only output the new or modified invariants you propose.
        \item No explanation. No commentary. Just show me the loop invariant.
    \end{enumerate}
\end{definedbox}
\vspace{-5mm}
\end{figure}
}

Since a single invariant property may require multiple propagation steps before all related obligations are discharged, \tool does not discard a candidate after the first failed repair.
Instead, we employ a delayed filtering policy to balance completeness and efficiency: if the number of refinement attempts for a given invariant exceeds the length of the longest cycle containing its node in the LCG, the candidate is deemed irrecoverable.
In such cases, further refinement is halted and the candidate is discarded, thereby pruning invalid propagation paths without prematurely rejecting invariants that require multi-step repair.

\section{Evaluation}\label{sec:exper}
In this section, we present an evaluation of \tool, which are designed to answer the following research questions.
\begin{description}
    \item[\textbf{RQ1.}] How does \tool perform compared with existing state-of-the-art loop invariant inference methods, especially for multi-loop programs?

    \item[\textbf{RQ2.}] What is the impact of different LLMs, obligation-guided inference, and WP-guided refinement strategy on \tool's performance?
\end{description}

\begin{table*}[!t]
\centering
\caption{Performance comparison of different loop invariant inference methods on the multi-loop benchmark. Numbers in parentheses denote the performance on the three  data sources: OOPSLA-13, SV-COMP, and CLRS-Alg. \tool solves at least 32 more problems, and performs especially better on the more challenging CLRS-Alg problems.}

\label{table:nested}
\begin{tabular}{lrrrrr}
\toprule
\multirow{2}{*}{Methods} & \multicolumn{2}{c}{\# Solved Benchmarks} &  \multirow{2}{*}{\# Avg. Proposals} & \multirow{2}{*}{Avg. Time (s)}  \\ 
\cmidrule(lr){2-3}
               & Only & Total 82 (16/8/58)             &               &                     \\
\midrule 
UAutomizer     &   1   & 28 (10/4/14)       &   3.2       &  12.8               \\
Code2Inv       &   0   & 10 (5/2/3)         &   6.2       &  69.5               \\
CLN2INV        &   0   & 15 (7/3/5)         &   7.3       &  2.7                \\
G-CLN          &   0   & 16 (7/3/6)         &   5.8       &  2.4                \\
LEMUR          &   0   & 29 (12/4/13)       &   1.3       &  70.9               \\
AutoSpec       &   1   & 40 (12/6/22)       &   1.2       &  134.6              \\
LaM4Inv        &   0   & 36 (14/5/17)       &   2.7       &  836.3              \\
\midrule
\tool          &   13  & \textbf{72 (16/8/48)} &   3.9    &  256.9              \\
\bottomrule
\end{tabular}
\end{table*}
 
\subsection{Experimental Setup}
\subsubsection{Benchmarks} 
To evaluate the performance of different loop invariant inference methods, we first include all existing loop-invariant benchmarks that contain multi-loop programs, including 46 problems from the OOPSLA-13 collection~\cite{dillig2013inductive} and 21 problems from SV-COMP~\cite{beyer2022progress}, both of which are widely used in prior work on loop invariant inference~\cite{wu2023lemur,wen2024enchanting,liu2025enhancing,yang2025automated}.
These 67 problems contain 24 programs with multiple loops and 43 programs with single loops, with an average of 1.6 loops per program.

To further assess performance on more realistic and challenging programs, 
we curate a new dataset \textbf{CLRS-Alg}, by manually formalizing 85 classic algorithms and exercises from Introduction to Algorithms (CLRS)~\cite{cormen2022introduction}. In contrast to the existing benchmarks that focused on numerical computations, CLRS-Alg includes 
algorithms commonly encountered in practice (e.g., sorting, searching, greedy algorithms, dynamic programming, and graph algorithms), and involves diverse data structures (e.g., arrays, queues, stacks, trees, graphs, and hash tables). 
As a result, these benchmarks 
require loop invariants that capture high-level semantic properties (e.g., array ordering or graph connectivity) rather than simple numerical constraints.
Among these 85 problems, 58 programs contain multiple loops and 27 contain single loops. On average, each CLRS-Alg program contains 2.6 loops.
Each problem is provided with its ACSL contracts and auxiliary predicates, while the evaluated methods are required to synthesize invariants for all loops so that the complete verification proof can be discharged. 

Overall, we partition the benchmarks into two categories based on loop complexity: (1) {\em single-loop benchmark} (70 problems) 
and (2) {\em multi-loop benchmark} (82 problems). 


\subsubsection{Baselines} We compare the following baselines.
\begin{itemize}[leftmargin=*]
    \item UAutomizer~\cite{heizmann2013software} is a model checker based on trace abstraction and predicate abstraction, utilizing an automata-theoretic approach for software verification.
    \item Code2Inv~\cite{si2018learning} employs a reinforcement learning framework with graph neural networks to capture program structures, learning to expand Backus-Naur Form rules for invariant generation.
    \item CLN2INV~\cite{ryan2019cln2inv} maps logical expressions to neural architectures, training continuous logic networks on execution traces to discover loop invariants.
    \item G-CLN~\cite{yao2020learning} extends CLN2INV by incorporating gating mechanisms and other enhancements, enabling the effective inference of complex nonlinear loop invariants.
    \item LEMUR~\cite{wu2023lemur} integrates LLMs with automated theorem provers, enhancing the $k$-induction capabilities of verifiers (e.g., ESBMC) by suggesting deductive steps and high-quality invariant candidates.
    \item AutoSpec~\cite{wen2024enchanting} combines static analysis with LLMs in an iterative framework, decomposing complex programs to guide the generation of satisfiable specifications. 
    \item LaM4Inv~\cite{wu2024llm} adopts a ``query-filter-reassemble'' paradigm, prompting LLMs to produce candidate invariants and utilizing formal verifiers to filter out incorrect ones iteratively via counterexamples.
\end{itemize}

Among these baselines, UAutomizer represents traditional formal methods that rely on symbolic reasoning. Code2Inv, CLN2INV, and G-CLN are learning-based approaches that focus on numerical programs by fitting logic to program features or execution traces. LEMUR, AutoSpec, and LaM4Inv are LLM-driven methods.

\subsubsection{Evaluation Metrics} 
To assess the performance of different loop invariant inference methods, we evaluate them using three metrics: 
(1) the number of benchmark instances for which valid loop invariants are successfully inferred; 
(2) the number of SMT solver invocations during the inference process; 
(3) the total inference time required to infer the loop invariant. 
The first metric measures the overall effectiveness, while the latter two metrics capture efficiency. Note that the efficiency metrics are computed only over the benchmark instances where a correct invariant is successfully generated.

\subsubsection{Implementation Details}
All methods are evaluated under identical experimental conditions. 
For LLMs used in \tool, we consider three LLMs: Qwen3.7-Max~\cite{qwen37}, DeepSeek-V4-Pro~\cite{xu2026deepseek}, and GPT-5.5~\cite{singh2025openai}.
%
For all LLM-based baselines,  GPT-5.5 is used as the default LLM. 
All LLMs use default settings (e.g., parameters {\em temperature} and {\em penalty} are both set to zero). 
We integrate three SMT solvers (i.e., Alt-Ergo~\cite{conchon2018alt}, Z3~\cite{de2008z3} and CVC5~\cite{barbosa2022cvc5}), and treat a query as successful as soon as any solver returns a definitive result. 
Each SMT solver invocation is limited to a timeout of 10 seconds. 
The WP reasoning is implemented using the WP plugin of the Frama-C framework (version 31.0, Gallium). 
All experiments are conducted on a GPU server equipped with an Intel Core i9-13900k@3.00GHz CPU, 48GB RAM, and a GeForce RTX 4090 GPU.


\subsection{RQ1. Effectiveness and Efficiency}

\begin{table*}[!t]
    \caption{Performance comparison of different loop invariant inference methods on the single-loop benchmark. Numbers in parentheses denote the performance on the three specific data sources: OOPSLA-13, SV-COMP, and CLRS-Alg. The proposed \tool solves 68 out of 70 problems, which is at least 19 more than the competitors.}
    \centering
    \label{table:single}
    \begin{tabular}{lrrrr} 
        \toprule
        \multirow{2}{*}{Methods} & \multicolumn{2}{c}{\# Solved Benchmarks} & \multirow{2}{*}{\# Avg. Proposals} & \multirow{2}{*}{Avg. Time (s)}  \\ 
        \cmidrule(lr){2-3}
                                 & Only & Total 70 (30/13/27)            &                                  &                       \\ 
        \midrule
        UAutomizer               &   0   & 35 (17/11/7)                   & 4.8                              & 9.2                   \\
        Code2Inv                 &   0   & 20 (13/5/2)                    & 9.7                             & 186.3 \\
        CLN2INV                  &   0   & 25 (14/8/3)                    & 47.3                             & 1.8                   \\
        G-CLN                    &   0   & 29 (15/10/4)                   & 46.8                             & 4.4                   \\
        LEMUR                    &   0   & 42 (25/10/7)                   & 1.6                              & 106.3                 \\
        AutoSpec                 &   1   & 49 (29/10/19)                  & 1.2                              & 32.9                  \\
        LaM4Inv                  &   0   & 47 (28/11/8)                   & 4.2                              & 47.6                  \\
        \midrule
        \tool                    &   6   & \textbf{68 (29/12/27)}         & 1.8                              & 24.9                  \\
        \bottomrule
    \end{tabular}
\end{table*}

\noindent{\bf (A) Results on multi-loop programs.} 
We initiate our evaluation by examining the generality of different invariant inference methods on the multi-loop benchmark. We compare \tool with the representative methods that are capable of reasoning over such complex control flows, namely, UAutomizer, LEMUR, AutoSpec and LaM4Inv. 
For baselines originally designed for single-loop programs, such as Code2Inv~\cite{si2018learning}, CLN2INV~\cite{ryan2019cln2inv}, and G-CLN~\cite{yao2020learning}, 
we extend their inference frameworks to process multi-loop programs so that they can be evaluated under the same benchmark setting. The results are summarized in Table~\ref{table:nested}. ``Only'' denotes the number of problems solved exclusively by the method, and ``Total'' denotes the total number of problems solved by that method.

Overall, \tool achieves the best performance, solving 72 out of the 82 benchmark problems, which is at least 32 more than the strongest competing method (i.e., AutoSpec). Moreover, \tool uniquely solves 13 problems that none of the existing competitors can solve. Notably, it demonstrates a substantial advantage on the more challenging CLRS-Alg problems. 
These results indicate that \tool is not limited to  numerical invariant (as in OOPSLA-13 and SV-COMP) but is capable of reasoning about 
complex algorithmic logic involving multiple loops (e.g., sortedness or connectivity). 
Although \tool incurs a higher average solving time due to its iterative refinement process, it maintains a low count of solver queries, demonstrating its efficiency in converging toward valid invariants despite the increased problem complexity. 

Among the baselines, the traditional tool UAutomizer performs poorly, solving only 28 instances. This result underscores the scalability limitations of predicate abstraction when applied to programs with complex control flows. As loop nesting depth increases, reasoning about array contents and heap-allocated data leads to severe state-space explosion and loss precision. 
The LLM-aided hybrid tools LEMUR and LaM4Inv exhibit similar limitations, solving 29 and 36 problems, respectively. 
Both approaches rely heavily on bounded model checking (specifically ESBMC) as the underlying verification engine. As a result, they inherit the fundamental limitations of bounded model checking when handling complex heap manipulations and quantified assertions in nested loop contexts, frequently failing to validate even syntactically correct invariants.
Code2Inv, CLN2INV, and G-CLN solve only 10, 15, and 16 problems, respectively. Their single-loop-oriented designs lack a mechanism to propagate proof contexts across loops, making them prone to weak invariants in the presence of \emph{local under-specification}.
AutoSpec achieves the second-best overall performance, solving 40 problems. However, its effectiveness is constrained by the absence of a constructive feedback mechanism. This limitation becomes especially pronounced on CLRS-Alg programs, where invariants typically require iterative refinement to capture inter-loop dependencies. 
%
Consequently, the tool struggles with verification tasks that require coordinated reasoning across complex programs.



We attribute the superior performance of \tool to the synergy between its obligation-guided inference strategy and WP-guided refinement. The framework explicitly propagates contextual information (such as preconditions and postconditions) from outer loops to inner ones. Moreover, the WP calculus serves as a principled feedback mechanism that propagates verification failures along the loop hierarchy, precisely localizing semantic mismatches. 
This enables the model to iteratively refine multiple related invariants in a synchronized and context-aware manner. 



\begin{table*}[!t]
    \caption{The ablation results of different components in \tool. Both our obligation-guided inference strategy and WP-based refinement play important roles.}
    \label{tab:ablation}
    \centering
    \begin{tabular}{lrrrrrrrrr} 
        \toprule
        \multirow{2}{*}{LLMs} & \multicolumn{3}{c}{\# Solved Benchmarks} & \multicolumn{3}{c}{\# Avg. Proposals} & \multicolumn{3}{c}{\# Avg. Time (s)} \\ 
        \cmidrule(lr){2-4}\cmidrule(lr){5-7}\cmidrule(lr){8-10}
        & No OGI & No WP & Full & No OGI & No WP & Full & No OGI & No WP & Full \\ 
        \midrule
        Qwen3.7-Max       & 93 & 96 & 112 & 1.1 & 1.1 & 1.3 & 68.7 & 67.5 & 133.9 \\
        DeepSeek-V4-Pro   & 80 & 76 & 117 & 1.5 & 1.2 & 2.2 & 73.8 & 58.4 & 350.4 \\
        GPT-5.5           & 117 & 110 & 140 & 3.4 & 1.1 & 2.9 & 74.8 & 36.2 & 144.2 \\
        \bottomrule
    \end{tabular}
\end{table*}

\smallskip 
\noindent{\bf (B) Results on single-loop programs.} To demonstrate the versatility and backward compatibility of our approach, we also evaluate \tool on single-loop programs. The results are presented in Table~\ref{table:single}, covering three benchmark sources: OOPSLA-13, SV-COMP, and CLRS-Alg. 

Overall, \tool still demonstrates superior effectiveness, successfully inferring valid loop invariants for 68 out of 70 benchmark problems. This surpasses the best-performing competitor, AutoSpec, by a margin of 19 solved problems. In terms of efficiency, \tool maintains a low average proposal count (1.8) with a moderate average solving time of 24.9 seconds. Moreover, \tool uniquely solves 6 single-loop problems that none of the existing competitors can solve.

Among the baselines, Code2Inv, CLN2INV and G-CLN 
show substantial degradation on the complex CLRS-Alg programs. The reason is twofold: (1) their underlying logic language for invariants are typically restricted to linear arithmetic, rendering them insufficient to express quantified properties (e.g., ``\textbackslash forall'', ``\textbackslash exists'') essential for verifying functional correctness in algorithms; and (2) their training data distributions largely exclude programs with complex data structures (e.g., array, stack and graph) present in CLRS-Alg.
Furthermore, consistent with the evaluation on multi-loops, traditional and hybrid tools (UAutomizer, LEMUR, LaM4Inv) continue to face bottlenecks imposed by their underlying verifiers (e.g., ESBMC) when applied to programs from the CLRS-Alg. 
AutoSpec performs relatively well on  single-loop CLRS-Alg programs. This is likely because the verification context is locally determined, where explicit pre- and post-conditions 
are available. 
However, this advantage does not extend to multi-loop settings, where iterative refinement across loops is essential.
In contrast, \tool exhibits strong generalizability driven by 
%
its constructive feedback mechanism, which enables the model to iteratively diagnose and correct errors. 

\smallskip
\noindent{\bf (C) Cross-dataset analysis}. Notably, a cross-dataset analysis reveals that the performance advantage of \tool widens significantly as the verification tasks become more challenging. While \tool maintains a solid lead on the numerical benchmarks (OOPSLA-13 and SV-COMP), solving 8 more instances than the runner-up AutoSpec, the disparity becomes far more pronounced on the rigorous CLRS-Alg dataset. On this benchmark with 85 problems, \tool surpasses AutoSpec by a significant margin of 34 solved problems (aggregating both single and multiple loops). This widening gap confirms that while existing LLM-based methods may suffice for simple control flows, \tool's hierarchical refinement capability is the deciding factor for conquering the high-level semantic complexity inherent in algorithmic verification.

\subsection{RQ2. Ablation Study}
To thoroughly evaluate the contribution of individual components in \tool, we conduct a series of ablation studies. These studies analyze performance variance across different LLMs, the obligation-guided inference strategy and the WP-guided refinement mechanism. Specifically, we consider three LLMs, including Qwen3.7-Max~\cite{qwen37}, DeepSeek-V4-Pro~\cite{xu2026deepseek}, and GPT-5.5~\cite{singh2025openai}.
The ``No OGI'' configuration replaces obligation-guided inference with a naive sequential inference order which ignores 
proof-obligation dependencies. 
The ``No WP'' configuration disables WP-guided failure diagnosis and dependency-propagating updates; when verification fails, the LLM only receives information about which invariant properties are violated. The ``Full'' configuration represents the complete implementation of \tool. The ablation results are summarized in Table~\ref{tab:ablation}.

\noindent{\bf (A) Impact of different LLMs}.
Table~\ref{tab:ablation} shows that \tool is effective across different backend LLMs. Even with Qwen3.7-Max and DeepSeek-V4-Pro, the ``Full'' configuration solves 112 and 117 benchmarks, respectively, demonstrating that our framework is not tied to a single proprietary model. With the stronger GPT-5.5 backend, \tool further improves to 140 solved benchmarks, indicating that the proposed inference and refinement mechanisms can exploit stronger reasoning capabilities when available. Overall, these results suggest that \tool provides robust gains across LLMs, and its performance scales with the reasoning capability of the underlying model.

\smallskip
\noindent{\bf (B) Impact of obligation-guided inference strategy}. In this ablation experiment, we compare the results of the ``No OGI'' configuration with those of the ``Full'' configuration. For instance, using DeepSeek-V4-Pro, the number of solved benchmarks rises from 80 to 117, while GPT-5.5 achieves an improvement from 117 to 140. Overall, these results demonstrate that obligation-guided inference provides critical proof contexts, enabling \tool to synthesize valid invariants for complex programs that cannot be handled when loops are treated in isolation.

\smallskip
\noindent{\bf (C) Impact of feedback information}.
Comparing the ``No WP'' and ``Full'' results underscores the impact of our refinement mechanism. The integration of WP-guided feedback yields substantial improvements in verification performance across all evaluated models. Specifically, for DeepSeek-V4-Pro, 
the number of solved benchmarks increases from 76 to 117. This demonstrates that the feedback effectively enables LLMs to rectify incorrect guesses. This mechanism is particularly indispensable for multiple loops, where the error space is vast. By leveraging WP calculus to identify specific violations and narrow the modification scope, our approach prevents the model from aimlessly exploring invalid predicates.  

\section{Discussion}\label{sec:discu}


We identify three primary threats to the validity of our study. First, our approach relies on the Frama-C platform and its WP plugin. However, the WP plugin faces significant challenges with dynamic memory management. As noted in the manual, it currently lacks native support for dynamic allocation~\cite{blanchard2018ghosts, blanchard2019logic}. Furthermore, Frama-C does not implement separation logic. Consequently, reasoning about pointer-based structures like linked lists requires expressing memory separation explicitly via verbose ACSL predicates and lemmas. This often results in complex proof obligations that overwhelm the underlying SMT solvers, leading to timeouts. Coq scripts may discharge some of these obligations, but automating such hybrid verification remains open.

Second, while our benchmark suite expands existing datasets by incorporating classical algorithms from Introduction to Algorithms,
it still falls short of representing the full diversity and complexity of real-world software. The selected programs, though nontrivial, remain relatively small and structured, limiting the generalizability of our findings to large-scale or system-level verification tasks. We aim to construct richer benchmarks featuring broader application domains in future work.

Third, there is a potential risk of data leakage in LLM-based evaluation. We cannot entirely exclude the possibility that benchmark programs from SV-COMP, OOPSLA, or textbook algorithms were included in the pretraining corpora of the underlying large language models. Nevertheless, we believe this risk is minimal for our task: while the source code or algorithmic descriptions may have appeared in public data, the precise formal specifications and ground-truth invariants used in our evaluation are manually annotated and not publicly available. Thus, the evaluation remains a fair and meaningful measure of the model’s reasoning ability.

\section{Related Work}\label{sec:relate}
This section reviews prior research on loop invariant inference, grouped into two major categories: traditional techniques and learning-based approaches.

\smallskip
\noindent{\bf Traditional approaches.}
Conventional research on invariant inference has primarily relied on static or dynamic program analysis.
Dynamic approaches, exemplified by Daikon~\cite{ernst2007daikon}, infer candidate invariants by observing variable behaviors during test executions.
Later extensions~\cite{sharma2013data,nguyen2014dig} integrate symbolic reasoning and dynamic traces to improve precision and soundness, though their success often depends heavily on the diversity and coverage of test inputs.
Moreover, their performance can degrade when handling a large or complex set of invariant templates.
Static approaches, in contrast, infer invariants directly from program semantics without requiring execution traces.
A wide variety of methods fall under this line of work, including logical abduction~\cite{calcagno2009bi,dillig2013inductive}, constraint solving~\cite{colon2003linear,gupta2009tests}, abstract interpretation~\cite{cousot1978automatic,cousot1979systematic,karr1976affine}, model checking~\cite{hojjat2018eldarica,vediramana2024global}, Craig interpolation~\cite{jhala2006practical,mcmillan2010lazy}, and syntax-guided synthesis (SyGuS)~\cite{barrett2011cvc4,barbosa2022cvc5}.
Tools such as LoopInvGen~\cite{padhi2016data}, which couples symbolic execution with constraint-based synthesis, and Eldarica~\cite{hojjat2018eldarica}, a Horn-clause–based model checker, are representative examples.
SMT solvers like CVC4 and CVC5~\cite{barrett2011cvc4,barbosa2022cvc5} have also been leveraged to synthesize invariants within the SyGuS framework.
Although these deductive methods are sound and precise, they typically face scalability and automation challenges, particularly when reasoning about non-linear arithmetic or heap-manipulating programs—thus motivating the exploration of learning-based paradigms.

\smallskip
\noindent{\bf Learning-based approaches.}
Recently, researchers have increasingly incorporated machine learning into invariant inference, typically following the “guess-and-check” strategy, in which candidate invariants are generated and then validated against program semantics.
Early studies adopt classic machine learning models such as decision trees~\cite{garg2014ice,garg2016learning,ezudheen2018horn,xu2020interval,riley2022multi}, support vector machines~\cite{sharma2012interpolants,li2017automatic}, and PAC learning~\cite{sharma2013verification}.
Later, reinforcement learning–based methods like Code2Inv~\cite{si2018learning} and LIPuS~\cite{yu2023loop} learn to iteratively expand invariant grammar trees based on feedback rewards.
Continuous logic–based approaches, including CLN2INV~\cite{ryan2019cln2inv} and G-CLN~\cite{yao2020learning}, further approximate symbolic reasoning via differentiable logical operators, enabling neural representations of invariant constraints.

The recent emergence of large language models (LLMs) has opened a new direction for invariant synthesis.
Kamath et al.~\cite{kamath2023finding} directly used pretrained LLMs to predict inductive invariants, while Chakraborty et al.~\cite{chakraborty2023ranking} proposed a ranking-based selection mechanism to filter plausible candidates.
Liu et al.~\cite{liu2024towards} introduced a self-supervised fine-tuning framework to better align LLMs with verification objectives, and Wen et al.~\cite{wen2024enchanting} decomposed verification tasks to guide the synthesis of specifications.
More recently, Wu et al.~\cite{wu2024llm} combined LLM with bounded model checking; Cao et al.~\cite{cao2025clause2inv} further extended it to non-linear cases, leveraging LLMs’ compositional generation capabilities for invariant clause synthesis; Yang et al.~\cite{yang2026integratingsymbolicexecutionllms} integrated symbolic execution to construct constraint templates that guide LLMs in instantiating valid loop invariants.
Different from these prior studies, our work focuses on the loop invariant inference problem of programs with complex loops. 

\section{Conclusion}\label{sec:concl}
In this paper, we present \tool, a neuro-symbolic framework designed to automate loop invariant inference for programs with complex loop structures. The core of \tool is to integrate an obligation-guided inference strategy with WP-guided refinement, built upon the loop-level abstraction. By introducing the loop-level call graph, \tool makes hierarchical loop relationships explicit, while WP calculus provides constructive feedback that enables coordinated invariant inference across multiple loops.
Experimental results on diverse benchmarks, including the newly curated CLRS-Alg dataset, demonstrate that \tool substantially outperforms state-of-the-art methods, especially on programs with multiple loops. 



\bibliographystyle{IEEEtran}
\bibliography{ref}

\begin{thebibliography}{10}
\providecommand{\url}[1]{#1}
\csname url@samestyle\endcsname
\providecommand{\newblock}{\relax}
\providecommand{\bibinfo}[2]{#2}
\providecommand{\BIBentrySTDinterwordspacing}{\spaceskip=0pt\relax}
\providecommand{\BIBentryALTinterwordstretchfactor}{4}
\providecommand{\BIBentryALTinterwordspacing}{\spaceskip=\fontdimen2\font plus
\BIBentryALTinterwordstretchfactor\fontdimen3\font minus
  \fontdimen4\font\relax}
\providecommand{\BIBforeignlanguage}[2]{{%
\expandafter\ifx\csname l@#1\endcsname\relax
\typeout{** WARNING: IEEEtran.bst: No hyphenation pattern has been}%
\typeout{** loaded for the language `#1'. Using the pattern for}%
\typeout{** the default language instead.}%
\else
\language=\csname l@#1\endcsname
\fi
#2}}
\providecommand{\BIBdecl}{\relax}
\BIBdecl

\bibitem{ezudheen2018horn}
P.~Ezudheen, D.~Neider, D.~D'Souza, P.~Garg, and P.~Madhusudan, ``Horn-ice
  learning for synthesizing invariants and contracts,'' \emph{Proceedings of
  the ACM on Programming Languages}, vol.~2, no. OOPSLA, pp. 1--25, 2018.

\bibitem{garg2016learning}
P.~Garg, D.~Neider, P.~Madhusudan, and D.~Roth, ``Learning invariants using
  decision trees and implication counterexamples,'' \emph{ACM Sigplan Notices},
  vol.~51, no.~1, pp. 499--512, 2016.

\bibitem{xu2020interval}
R.~Xu, F.~He, and B.-Y. Wang, ``Interval counterexamples for loop invariant
  learning,'' in \emph{Proceedings of the 28th ACM Joint Meeting on European
  Software Engineering Conference and Symposium on the Foundations of Software
  Engineering (ESEC/FSE)}, 2020, pp. 111--122.

\bibitem{si2018learning}
X.~Si, H.~Dai, M.~Raghothaman, M.~Naik, and L.~Song, ``Learning loop invariants
  for program verification,'' \emph{Advances in Neural Information Processing
  Systems (NeurIPS)}, vol.~31, 2018.

\bibitem{yu2023loop}
S.~Yu, T.~Wang, and J.~Wang, ``Loop invariant inference through smt solving
  enhanced reinforcement learning,'' in \emph{Proceedings of the 32nd ACM
  SIGSOFT International Symposium on Software Testing and Analysis (ISSTA)},
  2023, pp. 175--187.

\bibitem{ryan2019cln2inv}
G.~Ryan, J.~Wong, J.~Yao, R.~Gu, and S.~Jana, ``Cln2inv: Learning loop
  invariants with continuous logic networks,'' in \emph{International
  Conference on Learning Representations (ICLR)}, 2020.

\bibitem{yao2020learning}
J.~Yao, G.~Ryan, J.~Wong, S.~Jana, and R.~Gu, ``Learning nonlinear loop
  invariants with gated continuous logic networks,'' in \emph{Proceedings of
  the 41st ACM SIGPLAN Conference on Programming Language Design and
  Implementation (PLDI)}, 2020, pp. 106--120.

\bibitem{kamath2023finding}
A.~Kamath, A.~Senthilnathan, S.~Chakraborty, P.~Deligiannis, S.~K. Lahiri,
  A.~Lal, A.~Rastogi, S.~Roy, and R.~Sharma, ``Finding inductive loop
  invariants using large language models,'' \emph{arXiv preprint
  arXiv:2311.07948}, 2023.

\bibitem{chakraborty2023ranking}
\BIBentryALTinterwordspacing
S.~Chakraborty, S.~Lahiri, S.~Fakhoury, A.~Lal, M.~Musuvathi, A.~Rastogi,
  A.~Senthilnathan, R.~Sharma, and N.~Swamy, ``Ranking {LLM}-generated loop
  invariants for program verification,'' in \emph{Findings of the Association
  for Computational Linguistics: EMNLP 2023}, H.~Bouamor, J.~Pino, and K.~Bali,
  Eds.\hskip 1em plus 0.5em minus 0.4em\relax Singapore: Association for
  Computational Linguistics, Dec. 2023, pp. 9164--9175. [Online]. Available:
  \url{https://aclanthology.org/2023.findings-emnlp.614/}
\BIBentrySTDinterwordspacing

\bibitem{liu2024towards}
C.~Liu, X.~Wu, Y.~Feng, Q.~Cao, and J.~Yan, ``Towards general loop invariant
  generation: a benchmark of programs with memory manipulation,''
  \emph{Advances in Neural Information Processing Systems}, vol.~37, pp.
  129\,120--129\,145, 2024.

\bibitem{wu2023lemur}
H.~Wu, C.~Barrett, and N.~Narodytska, ``Lemur: Integrating large language
  models in automated program verification,'' in \emph{The Twelfth
  International Conference on Learning Representations (ICLR)}, 2024.

\bibitem{beyer2022progress}
D.~Beyer, ``Progress on software verification: Sv-comp 2022,'' in
  \emph{International Conference on Tools and Algorithms for the Construction
  and Analysis of Systems}.\hskip 1em plus 0.5em minus 0.4em\relax Springer,
  2022, pp. 375--402.

\bibitem{dillig2013inductive}
\BIBentryALTinterwordspacing
I.~Dillig, T.~Dillig, B.~Li, and K.~McMillan, ``Inductive invariant generation
  via abductive inference,'' \emph{SIGPLAN Not.}, vol.~48, no.~10, p.
  443–456, Oct. 2013. [Online]. Available:
  \url{https://doi.org/10.1145/2544173.2509511}
\BIBentrySTDinterwordspacing

\bibitem{cormen2022introduction}
T.~H. Cormen, C.~E. Leiserson, R.~L. Rivest, and C.~Stein, \emph{Introduction
  to algorithms}.\hskip 1em plus 0.5em minus 0.4em\relax MIT press, 2022.

\bibitem{heizmann2013software}
M.~Heizmann, J.~Hoenicke, and A.~Podelski, ``Software model checking for people
  who love automata,'' in \emph{International Conference on Computer Aided
  Verification}.\hskip 1em plus 0.5em minus 0.4em\relax Springer, 2013, pp.
  36--52.

\bibitem{wu2024llm}
G.~Wu, W.~Cao, Y.~Yao, H.~Wei, T.~Chen, and X.~Ma, ``Llm meets bounded model
  checking: Neuro-symbolic loop invariant inference,'' in \emph{Proceedings of
  the 39th IEEE/ACM International Conference on Automated Software
  Engineering}, 2024, pp. 406--417.

\bibitem{wen2024enchanting}
C.~Wen, J.~Cao, J.~Su, Z.~Xu, S.~Qin, M.~He, H.~Li, S.-C. Cheung, and C.~Tian,
  ``Enchanting program specification synthesis by large language models using
  static analysis and program verification,'' in \emph{International Conference
  on Computer Aided Verification}.\hskip 1em plus 0.5em minus 0.4em\relax
  Springer, 2024, pp. 302--328.

\bibitem{de2008z3}
L.~De~Moura and N.~Bj{\o}rner, ``Z3: An efficient smt solver,'' in
  \emph{International conference on Tools and Algorithms for the Construction
  and Analysis of Systems}.\hskip 1em plus 0.5em minus 0.4em\relax Springer,
  2008, pp. 337--340.

\bibitem{dijkstra1976discipline}
E.~W. Dijkstra, \emph{A discipline of programming}.\hskip 1em plus 0.5em minus
  0.4em\relax prentice-hall Englewood Cliffs, 1976, vol. 613924118.

\bibitem{cuoq2012frama}
P.~Cuoq, F.~Kirchner, N.~Kosmatov, V.~Prevosto, J.~Signoles, and B.~Yakobowski,
  ``Frama-c: A software analysis perspective,'' in \emph{International
  conference on software engineering and formal methods}.\hskip 1em plus 0.5em
  minus 0.4em\relax Springer, 2012, pp. 233--247.

\bibitem{hatcliff2012behavioral}
J.~Hatcliff, G.~T. Leavens, K.~R.~M. Leino, P.~M{\"u}ller, and M.~Parkinson,
  ``Behavioral interface specification languages,'' \emph{ACM Computing Surveys
  (CSUR)}, vol.~44, no.~3, pp. 1--58, 2012.

\bibitem{liu2025enhancing}
R.~Liu, M.~Chen, L.-I. Wu, J.~Ke, and G.~Li, ``Enhancing automated loop
  invariant generation for complex programs with large language models,''
  \emph{Science of Computer Programming}, p. 103387, 2025.

\bibitem{yang2025automated}
F.~Yang, X.~Ma, S.~Wang, X.~Xu, Q.~Cao, N.~Zhan, X.~Li, and B.~Gu, ``Automated
  synthesis of formally verified multi-abstraction function summaries,''
  \emph{arXiv preprint arXiv:2506.09550}, 2025.

\bibitem{qwen37}
\BIBentryALTinterwordspacing
{Qwen Team}, ``{Qwen3.7}: The agent frontier,'' May 2026. [Online]. Available:
  \url{https://qwen.ai/blog?id=qwen3.7}
\BIBentrySTDinterwordspacing

\bibitem{xu2026deepseek}
A.~Xu, B.~Lin, B.~Xue, B.~Wang, B.~Xu, B.~Wu, B.~Zhang, C.~Lin, C.~Dong,
  C.~Ling \emph{et~al.}, ``Deepseek-v4: Towards highly efficient million-token
  context intelligence,'' \emph{arXiv preprint arXiv:2606.19348}, 2026.

\bibitem{singh2025openai}
A.~Singh, A.~Fry, A.~Perelman, A.~Tart, A.~Ganesh, A.~El-Kishky, A.~McLaughlin,
  A.~Low, A.~Ostrow, A.~Ananthram \emph{et~al.}, ``Openai gpt-5 system card,''
  \emph{arXiv preprint arXiv:2601.03267}, 2025.

\bibitem{conchon2018alt}
S.~Conchon, A.~Coquereau, M.~Iguernlala, and A.~Mebsout, ``Alt-ergo 2.2,'' in
  \emph{SMT Workshop: International Workshop on Satisfiability Modulo
  Theories}, 2018.

\bibitem{barbosa2022cvc5}
H.~Barbosa, C.~Barrett, M.~Brain, G.~Kremer, H.~Lachnitt, M.~Mann, A.~Mohamed,
  M.~Mohamed, A.~Niemetz, A.~N{\"o}tzli \emph{et~al.}, ``cvc5: A versatile and
  industrial-strength smt solver,'' in \emph{International Conference on Tools
  and Algorithms for the Construction and Analysis of Systems}.\hskip 1em plus
  0.5em minus 0.4em\relax Springer, 2022, pp. 415--442.

\bibitem{blanchard2018ghosts}
A.~Blanchard, N.~Kosmatov, and F.~Loulergue, ``Ghosts for lists: a critical
  module of contiki verified in frama-c,'' in \emph{NASA Formal Methods
  Symposium}.\hskip 1em plus 0.5em minus 0.4em\relax Springer, 2018, pp.
  37--53.

\bibitem{blanchard2019logic}
------, ``Logic against ghosts: Comparison of two proof approaches for a list
  module,'' in \emph{Proceedings of the 34th ACM/SIGAPP Symposium on Applied
  Computing}, 2019, pp. 2186--2195.

\bibitem{ernst2007daikon}
M.~D. Ernst, J.~H. Perkins, P.~J. Guo, S.~McCamant, C.~Pacheco, M.~S. Tschantz,
  and C.~Xiao, ``The daikon system for dynamic detection of likely
  invariants,'' \emph{Science of computer programming}, vol.~69, no. 1-3, pp.
  35--45, 2007.

\bibitem{sharma2013data}
R.~Sharma, S.~Gupta, B.~Hariharan, A.~Aiken, P.~Liang, and A.~V. Nori, ``A data
  driven approach for algebraic loop invariants,'' in \emph{European Symposium
  on Programming}.\hskip 1em plus 0.5em minus 0.4em\relax Springer, 2013, pp.
  574--592.

\bibitem{nguyen2014dig}
T.~Nguyen, D.~Kapur, W.~Weimer, and S.~Forrest, ``Dig: A dynamic invariant
  generator for polynomial and array invariants,'' \emph{ACM Transactions on
  Software Engineering and Methodology (TOSEM)}, vol.~23, no.~4, pp. 1--30,
  2014.

\bibitem{calcagno2009bi}
\BIBentryALTinterwordspacing
C.~Calcagno, D.~Distefano, and V.~Vafeiadis, ``Bi-abductive resource invariant
  synthesis,'' in \emph{Proceedings of the 7th Asian Symposium on Programming
  Languages and Systems}, ser. APLAS '09.\hskip 1em plus 0.5em minus
  0.4em\relax Berlin, Heidelberg: Springer-Verlag, 2009, p. 259–274.
  [Online]. Available: \url{https://doi.org/10.1007/978-3-642-10672-9_19}
\BIBentrySTDinterwordspacing

\bibitem{colon2003linear}
M.~A. Col{\'o}n, S.~Sankaranarayanan, and H.~B. Sipma, ``Linear invariant
  generation using non-linear constraint solving,'' in \emph{International
  Conference on Computer Aided Verification (CAV)}.\hskip 1em plus 0.5em minus
  0.4em\relax Springer, 2003, pp. 420--432.

\bibitem{gupta2009tests}
A.~Gupta, R.~Majumdar, and A.~Rybalchenko, ``From tests to proofs,'' in
  \emph{International Conference on Tools and Algorithms for the Construction
  and Analysis of Systems}.\hskip 1em plus 0.5em minus 0.4em\relax Springer,
  2009, pp. 262--276.

\bibitem{cousot1978automatic}
\BIBentryALTinterwordspacing
P.~Cousot and N.~Halbwachs, ``Automatic discovery of linear restraints among
  variables of a program,'' in \emph{Proceedings of the 5th ACM SIGACT-SIGPLAN
  Symposium on Principles of Programming Languages}, ser. POPL '78.\hskip 1em
  plus 0.5em minus 0.4em\relax New York, NY, USA: Association for Computing
  Machinery, 1978, p. 84–96. [Online]. Available:
  \url{https://doi.org/10.1145/512760.512770}
\BIBentrySTDinterwordspacing

\bibitem{cousot1979systematic}
\BIBentryALTinterwordspacing
P.~Cousot and R.~Cousot, ``Systematic design of program analysis frameworks,''
  in \emph{Proceedings of the 6th ACM SIGACT-SIGPLAN Symposium on Principles of
  Programming Languages}, ser. POPL '79.\hskip 1em plus 0.5em minus 0.4em\relax
  New York, NY, USA: Association for Computing Machinery, 1979, p. 269–282.
  [Online]. Available: \url{https://doi.org/10.1145/567752.567778}
\BIBentrySTDinterwordspacing

\bibitem{karr1976affine}
M.~Karr, ``Affine relationships among variables of a program,'' \emph{Acta
  Informatica}, vol.~6, no.~2, pp. 133--151, 1976.

\bibitem{hojjat2018eldarica}
H.~Hojjat and P.~R{\"u}mmer, ``The eldarica horn solver,'' in \emph{2018 Formal
  Methods in Computer Aided Design (FMCAD)}.\hskip 1em plus 0.5em minus
  0.4em\relax IEEE, 2018, pp. 1--7.

\bibitem{vediramana2024global}
H.~G. Vediramana~Krishnan, Y.~Chen, S.~Shoham, and A.~Gurfinkel, ``Global
  guidance for local generalization in model checking,'' \emph{Formal Methods
  in System Design}, vol.~63, no.~1, pp. 81--109, 2024.

\bibitem{jhala2006practical}
\BIBentryALTinterwordspacing
R.~Jhala and K.~L. McMillan, ``A practical and complete approach to predicate
  refinement,'' in \emph{Proceedings of the 12th International Conference on
  Tools and Algorithms for the Construction and Analysis of Systems}, ser.
  TACAS'06.\hskip 1em plus 0.5em minus 0.4em\relax Berlin, Heidelberg:
  Springer-Verlag, 2006, p. 459–473. [Online]. Available:
  \url{https://doi.org/10.1007/11691372_33}
\BIBentrySTDinterwordspacing

\bibitem{mcmillan2010lazy}
\BIBentryALTinterwordspacing
K.~L. McMillan, ``Lazy annotation for program testing and verification,'' in
  \emph{Proceedings of the 22nd International Conference on Computer Aided
  Verification}, ser. CAV'10.\hskip 1em plus 0.5em minus 0.4em\relax Berlin,
  Heidelberg: Springer-Verlag, 2010, p. 104–118. [Online]. Available:
  \url{https://doi.org/10.1007/978-3-642-14295-6_10}
\BIBentrySTDinterwordspacing

\bibitem{barrett2011cvc4}
C.~Barrett, C.~L. Conway, M.~Deters, L.~Hadarean, D.~Jovanovi\'{c}, T.~King,
  A.~Reynolds, and C.~Tinelli, ``Cvc4,'' in \emph{Proceedings of the 23rd
  International Conference on Computer Aided Verification}, ser. CAV'11.\hskip
  1em plus 0.5em minus 0.4em\relax Berlin, Heidelberg: Springer-Verlag, 2011,
  p. 171–177.

\bibitem{padhi2016data}
S.~Padhi, R.~Sharma, and T.~Millstein, ``Data-driven precondition inference
  with learned features,'' \emph{ACM SIGPLAN Notices}, vol.~51, no.~6, pp.
  42--56, 2016.

\bibitem{garg2014ice}
P.~Garg, C.~L{\"o}ding, P.~Madhusudan, and D.~Neider, ``Ice: A robust framework
  for learning invariants,'' in \emph{International Conference on Computer
  Aided Verification (CAV)}.\hskip 1em plus 0.5em minus 0.4em\relax Springer,
  2014, pp. 69--87.

\bibitem{riley2022multi}
\BIBentryALTinterwordspacing
D.~Riley and G.~Fedyukovich, ``Multi-phase invariant synthesis,'' in
  \emph{Proceedings of the 30th ACM Joint European Software Engineering
  Conference and Symposium on the Foundations of Software Engineering}, ser.
  ESEC/FSE 2022.\hskip 1em plus 0.5em minus 0.4em\relax New York, NY, USA:
  Association for Computing Machinery, 2022, p. 607–619. [Online]. Available:
  \url{https://doi.org/10.1145/3540250.3549166}
\BIBentrySTDinterwordspacing

\bibitem{sharma2012interpolants}
\BIBentryALTinterwordspacing
R.~Sharma, A.~V. Nori, and A.~Aiken, ``Interpolants as classifiers,'' in
  \emph{Proceedings of the 24th International Conference on Computer Aided
  Verification}, ser. CAV'12.\hskip 1em plus 0.5em minus 0.4em\relax Berlin,
  Heidelberg: Springer-Verlag, 2012, p. 71–87. [Online]. Available:
  \url{https://doi.org/10.1007/978-3-642-31424-7_11}
\BIBentrySTDinterwordspacing

\bibitem{li2017automatic}
J.~Li, J.~Sun, L.~Li, Q.~L. Le, and S.-W. Lin, ``Automatic loop-invariant
  generation anc refinement through selective sampling,'' in \emph{2017 32nd
  IEEE/ACM International Conference on Automated Software Engineering (ASE)},
  2017, pp. 782--792.

\bibitem{sharma2013verification}
R.~Sharma, S.~Gupta, B.~Hariharan, A.~Aiken, and A.~V. Nori, ``Verification as
  learning geometric concepts,'' in \emph{Static Analysis: 20th International
  Symposium, SAS 2013, Seattle, WA, USA, June 20-22, 2013. Proceedings
  20}.\hskip 1em plus 0.5em minus 0.4em\relax Springer, 2013, pp. 388--411.

\bibitem{cao2025clause2inv}
W.~Cao, G.~Wu, T.~Xu, Y.~Yao, H.~Wei, T.~Chen, and X.~Ma, ``Clause2inv: A
  generate-combine-check framework for loop invariant inference,''
  \emph{Proceedings of the ACM on Software Engineering}, vol.~2, no. ISSTA, pp.
  1009--1030, 2025.

\bibitem{yang2026integratingsymbolicexecutionllms}
\BIBentryALTinterwordspacing
F.~Yang, X.~Ma, S.~Wang, X.~Xu, Q.~Cao, N.~Zhan, X.~Li, and B.~Gu,
  ``Integrating symbolic execution with llms for automated generation of
  program specifications,'' 2026. [Online]. Available:
  \url{https://arxiv.org/abs/2506.09550}
\BIBentrySTDinterwordspacing

\end{thebibliography}
\end{document}